\documentclass[letterpaper, 10 pt, conference]{ieeeconf}  %

\IEEEoverridecommandlockouts                              %

\usepackage[backend=bibtex,defernumbers=true,sorting=none,style=ieee]{biblatex}%
\usepackage{microtype}
\usepackage{epsfig} %
\usepackage{newtxtext}
\usepackage{newtxmath}
\usepackage{times} %
\usepackage{amsmath} %
\usepackage{wasysym} %

\usepackage{graphicx}%
\usepackage{multirow}%
\usepackage[none]{hyphenat}%
\usepackage{float}%
\usepackage{subfig}%
\usepackage{iftex}%
\usepackage{textcomp}

\usepackage{hyperref} 
\usepackage{soul}
\usepackage[dvipsnames]{xcolor}
\usepackage{textcomp}
\usepackage{marvosym}

\usepackage{siunitx}
\makeatletter\@ifundefined{qty}{%
  \def\qty{\SI}%
  \def\qtylist{\SIlist}%
}\makeatother

\usepackage[normalem]{ulem}

\definecolor{twred}{rgb}{0.85,0,0.2}
\definecolor{twgreen}{rgb}{0,0.65,0.2}

\def\clap#1{\hbox to 0pt{\hss #1\hss}}%
\def\initials#1{\protect\clap{\smash{\raisebox{1.4ex}{\tiny{\textsf{\textit{~~#1}}}}}}}%
\makeatletter
\newcommand{\NOTE}[3]{\protect\@ifundefined{hidecomments}{%
  \strut{\color{#2}{\hspace{0pt}\initials{#1}\protect{{\small$\lfloor$}#3{\small]}}}}%
  }{}}
\newcommand{\EDITbyauthor}[4][]{\protect\@ifundefined{hidecomments}{%
  \strut{\color{#3}{\hspace{0pt}\initials{#2}\protect\sout{#1}{#4}}}%
  }{}}
\newcommand{\EDITredandgreen}[4][]{\protect\@ifundefined{hidecomments}{%
  \strut{\color{twred}{\protect\sout{#1}{\color{twgreen}{#4}}}}%
  }{}}
\newcommand{\EDITgreenonly}[4][]{\protect\@ifundefined{hidecomments}{%
  \strut{\color{twgreen}{#4}}%
  }{}}
\newcommand{\EDITfinal}[4][]{\protect%
  \strut{#4}%
  {}}

\newcommand{\EDIT}[4][]{\EDITfinal[#1]{#2}{#3}{#4}}

\newcommand{\NOTEboxed}[3]{\protect\@ifundefined{hidecomments}{%
  {\centering\fbox{\parbox{0.97\linewidth}{\protect\EDIT{#1}{#2}{#3}}}}%
  }{}}

\def\myhrulefill{\leavevmode\leaders\hrule height 2pt\hfill\kern\z@}

\makeatother

\newcommand{\VWedit}[2][]{\protect\EDIT[#1]{VW}{Orange}{#2}}

\newcommand{\TWedit}[2][]{\protect\EDIT[#1]{TW}{twgreen}{#2}}

\def\ignore#1{}

\usepackage{bm}

\newcommand{\vect}[1]{{\ensuremath{\boldsymbol{#1}}}}    %

\title{\LARGE \bf
Automatic Spatial Calibration of Near-Field MIMO Radar\\  With Respect to Optical \VWedit{Depth }Sensors
}

\author{
	Vanessa Wirth \textsuperscript{\Letter}$^{,1}$\qquad
	Johanna Br{\"a}unig$^{2}$\qquad
	Danti Khouri$^{2}$\qquad
	Florian Gutsche$^{1}$\\[0.75ex] 
	Martin Vossiek$^{2}$\qquad
	Tim Weyrich$^{\ast,1,3}$\qquad
	Marc Stamminger$^{\ast,1}$
\thanks{$\ast$\mbox{~}These authors contributed equally to this work}%
\thanks{$^{1}$\mbox{~}Visual Computing Erlangen (VCE),
	Friedrich-Alexander-Universit{\"a}t
	\mbox{\qquad\quad}~Erlangen-N{\"u}rnberg, Germany}%
\thanks{$^{2}$\mbox{~}Institute of Microwaves and Photonics (LHFT), Friedrich-Alexander- \mbox{\qquad\quad}Universit{\"a}t~Erlangen-N{\"u}rnberg, Germany}%
\thanks{$^{3}$\mbox{~}Dept of Computer Science, University College London, UK}%
\thanks{\Letter \href{mailto:vanessa.wirth@fau.de}{\nolinkurl{vanessa.wirth@fau.de}}}%
}

\begin{document}

\maketitle
\thispagestyle{empty}
\pagestyle{empty}

\begin{abstract}
  Despite an emerging interest in MIMO radar, the utilization of its
  complementary strengths in combination with optical \VWedit{depth }sensors has so
  far been limited to far-field applications, due to the challenges
  that arise from mutual sensor calibration in the near field.
  In fact, most related approaches in the autonomous industry propose
  target-based calibration methods using corner reflectors
  that have proven to be unsuitable for the near field.
  In contrast, we propose a novel, joint calibration approach for
  optical RGB-D sensors and MIMO radars that is designed to operate in
  the radar's \emph{near-field range}, within decimeters from the
  sensors.
  Our pipeline consists of a bespoke calibration target, allowing for
  automatic target detection and localization, followed by the spatial
  calibration of the two sensor coordinate systems through target
  registration.
  We validate our approach using two different depth sensing
  technologies from the optical domain.
  The experiments show the efficiency and accuracy of our
  calibration for various target displacements, as well as its
  robustness of our localization in terms of signal ambiguities.
\end{abstract}

\section{Introduction}

The ability to sense an environment in terms of accurate 3D information is crucial for many applications, including robotics, autonomous driving, or human-computer interaction.
  A prominent sensor class \VWedit[are ]{is }range-sensing imagers; this work considers both optical imagers as well as imaging radar.

Driven by data availability, high spatial resolution, and low cost, optical depth sensing technologies such as time-of-flight cameras and single- or multi-view stereo algorithms
are widely used;
a
tremendous amount of research has been conducted, for example, in the field of static~\cite{zollhoefer_2018} and dynamic~\cite{xian_feng_2021, zollhoefer_2018, tretschk_2023} reconstruction, human pose and shape estimation~\cite{yian_2023}, and scene understanding~\cite{naseer_2019}.

\begin{figure}[t]
	\centering\vspace*{-1ex}
	\includegraphics[width=0.95\linewidth]{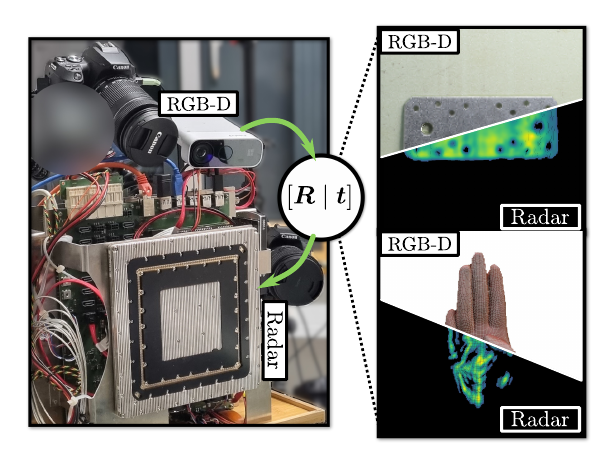}\\
	\caption{Our calibration estimates the relative rotation $R$ and translation $t$ between an optical RGB-D sensor and an imaging MIMO radar incorporating high angular resolution. %
	}
	\label{fig:teaser}
\end{figure}

On the other hand, a growing interest has emerged with respect to radar imaging, prominently utilized for security scanning~\cite{sherif_2012, sherif_2021} and autonomous driving~\cite{schwarz_2022, bialer_2021}.
Radar is able to provide range cues in the presence of fog or dust, can penetrate fabric, and is insensitive to environmental light. 
Compared to camera-based systems, radar imaging is a recent range-sensing technology that involves calculating spatial object or feature distributions, commonly by using digital beamforming.
Popular sensors are multiple-input multiple-output (MIMO) radars, which process the received signals coherently to form a synthetic antenna aperture by comparing the phase difference between multiple incoming signals at distinctive spatial receiver positions.
They essentially exploit that each antenna (both transmitters and receivers) looks at scene points from different directions, which allows to 3D reconstruct a scene from the resulting phase differences.
The result is commonly represented as a voxel grid or point cloud (cf. \autoref{fig:teaser}), including confidence values about a target's presence that are proportional to the received signal power.

\begin{figure*}[t]
	\vspace*{-1ex}\centering
	\includegraphics[width=0.95\linewidth]{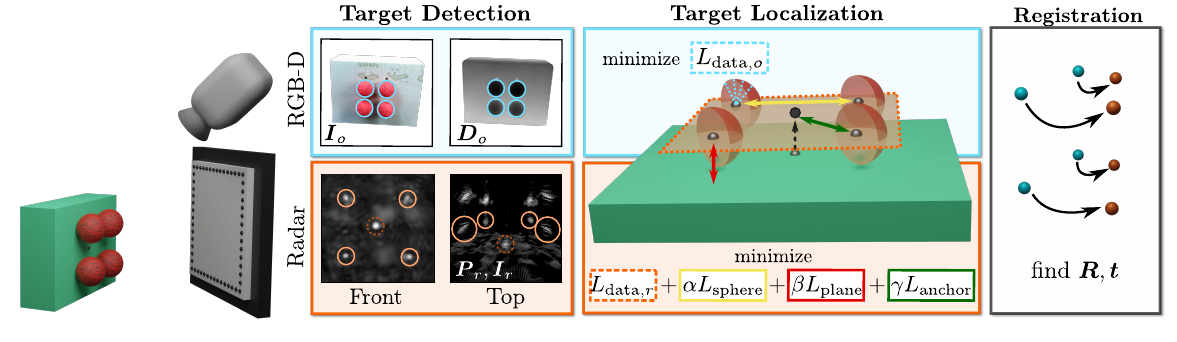}\\
	\caption{\label{fig:overview}%
          The calibration is divided into sensor-specific parts for
          target detection and target localization. To acquire the
          calibration parameters, we register the localized target
          points from the optical domain (blue) to points of the radar
          domain (orange).\vspace*{-1ex}}
\end{figure*}

A growing body of work~\cite{ heng_2020, chen_2022,domhof_2019, lee_2020, cheng_2023, lee_2023, shiva_2023, choi_2023} recognizes the potential of combining optical depth sensors (which we collectively refer to as \emph{RGB-D sensors}), and MIMO radars. However, they invariably operate in the radar's \emph{far field}\TWedit[, that is]{}, at distances where standard solutions exist to mutually calibrate (align) the\TWedit[ir]{} respective sensor coordinate systems.
In contrast, we take on
the unique challenge of localizing joint calibration objects within the MIMO radar's\,\emph{near-field range}, i.e.,\,within few decimeters, where traditional radar targets appear \mbox{strongly distorted}.

While a small number of previous works in the context of autonomous driving compute the \VWedit{spatial }calibration on the fly during the capture process, e.g., by leveraging motion cues\VWedit[~\cite{heng_2020}]{~\cite{heng_2020, wise_2023}} during a car drive, most calibrations are static and target-based, that is, a specific calibration target~\cite{chen_2022,domhof_2019, lee_2020, cheng_2023, lee_2023, shiva_2023, choi_2023} is designed to yield robust and accurate reconstruction results in all relevant sensors. 
The primary target of choice to be detected by a radar in the far field is a metal, trihedral corner reflector~\cite{domhof_2019, lee_2020, cheng_2023, lee_2023, shiva_2023, choi_2023} because of its strong echo signal for a comparatively large range of acceptance angles:
in far-field conditions, the signal propagation can be approximated as parallel to the radar's line of sight, resulting in a retroreflective behavior for various antenna positions.
Moreover, the reflector geometry ensures a total path length of the received signal that is constant across its entire aperture, which is why the corner is reconstructed as a bright, seemingly planar reflector that can easily be detected automatically.

\begin{figure}[b]
	\centering\vspace*{-1ex}  %
	\includegraphics[width=\linewidth]{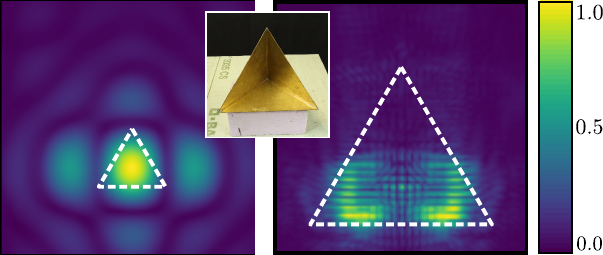}
	\caption{\label{fig:corner}%
		Target confidence of a corner reflector captured by a MIMO radar at \qty{2.6}{\meter} (\textit{left}) and \qty{0.3}{\meter} (\textit{right}) distance.\vspace*{-0.8ex}}
\end{figure}

For MIMO arrays in near-field scenarios with large angles between a target and a transmitter-receiver antenna pair, however, the desired properties of a corner reflector do not hold~\cite{liu_2017}.
\autoref{fig:corner} illustrates the distinctive signal response between a corner reflector captured in the far field and in the near field.
For this reason, we conclude that the aforementioned line of related work is not suitable for calibration in the near field, where the target has only a few decimeters distance to the sensor.
An orthogonal approach to ours proposed by Chen et al.~\cite{chen_2022} leverages optical markers as target to calibrate an imaging radar with a motion capture system for reconstruction of human bodies. 
The method introduced in this paper can be applied to optical RGB-D technologies with a 2D-3D correspondence relationship, for example time-of-flight and single- or multi-view stereo systems.
These systems provide 2D depth maps, which can be back-projected into 3D given the camera's intrinsic parameters. 
To the best of our knowledge, we are the first to propose an automatic target-based calibration method for optical RGB-D sensors and imaging MIMO radars in the near field.

To achieve this, the contributions of this paper are:
\begin{itemize}
	\item \TWedit[The design ]{Design }of a calibration target that is robustly detectable from various optical RGB-D\TWedit[ technologies]{ sensors} and MIMO radars.
	\item An automatic pipeline for target detection and localization, followed by a spatial registration.
	\item An overall framework that yields precise calibration parameters with millimeter accuracy, assessed by pairing a MIMO radar sensor with two different RGB-D technologies, time-of-flight and multi-view stereo.
\end{itemize}

\subsection{Overview}
Our full pipeline is illustrated in~\autoref{fig:overview}.
We capture a calibration target, which is specifically designed for near-field conditions, from an optical RGB-D sensor and a MIMO radar.
The optical sensor provides an RGB and a depth image, denoted as $\vect{I}_o\in \mathbb{R}^{W\times H \times 3}$ and $\vect{D}_o\in \mathbb{R}^{W\times H}$, respectively.
Furthermore, we acquire a radar point cloud $\vect{P}_r \in \mathbb{R}^{N \times 3}$ with confidence values proportional to the received signal amplitude $\vect{I}_r \in [0,1]^N$.
Since the visibility of materials and geometries depends on the received signal wavelength, the target detection as well as the target localization are divided into sensor-specific parts.
During detection, our method finds possible target candidates.
Given these candidates, the localization stage utilizes sensor-specific prior knowledge about the calibration target to filter outliers and calculate the spatial position of the point samples that are used for the following registration stage.
During registration, our method computes the optimal transformation between point samples from the optical and the radar domain, respectively.
Lastly, for evaluation, we optionally employ an additional refinement stage with a second capture target.

\begin{figure}[]
	\centering\vspace*{-1.5ex}%
	\includegraphics[width=1\linewidth]{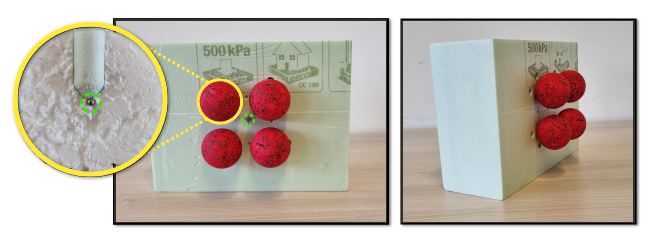}\\
	\caption{\label{fig:target}%
		The calibration target consists of four styrofoam spheres (\diameter~\qty{5}{\cm}), each with a steel ball (\diameter~\qty{2.5}{\mm}) embedded at its center; sphere centers form a square of \qty{6}{\cm} edge length. A fifth steel ball is centered on the styrofoam back plane.\vspace*{-0.5ex}}
	
\end{figure}

\section{Calibration Target}
To establish correspondences between an optical depth sensor and a MIMO radar, a calibration target is required that is robustly detectable, despite the significant domain gap between the different operating wavelengths. 
We opted for a target that can be detected within a wider range of viewing angles, to avoid having to precisely align the target in front the MIMO radar, as would be required for many potential target geometries where reconstruction quality significantly depends on the angle of incidence to the transmitters.
Our calibration target is depicted in \autoref{fig:target} and consists of four textured styrofoam spheres, arranged in a square and mounted onto a styrofoam board. 
While styrofoam is a material hardly visible to radar, the spheres contain smaller, highly radar-reflecting steel balls inside\VWedit{, which were embedded using a high-precision drill}.
An additional steel ball at the center of the square is placed onto the styrofoam board.
We will now elaborate
on our design choices. 
First, we chose \VWedit{view-independent }spherical shapes since hard corners and edges are challenging to detect in range sensors\VWedit{ with millimeter accuracy due to multi-path signal interference at object silhouettes~\cite{zanuttigh_2016}}. 
Furthermore, due to its material properties, metal is highly reflective for radar signals but sharp edges lead to diffraction, which introduces sidelobes and other types of noise into the reconstructed point cloud. 
Therefore, we took countermeasures to ensure the signal-to-noise ratio of our target is as high as possible. 
The square arrangement, which shares the symmetry \VWedit{and, approximately, the spatial extents }of the MIMO antenna layout, \VWedit{provides a calibration feature distribution (see~\autoref{sec:detection}) within the maximum focused area of the radar field of view, }helps balancing out symmetric noise artifacts and
enables the possibility of outlier detection through spatial constraints.
Ensuring uniformity in the spot visible to each transmitting antenna, similar to the assumption made for corner reflectors in the far field, the diameter of the metal spheres is chosen such that the reconstructed signal is close to a single point target:
in our setup, the diameter of \qty{2.5}{\mm} is smaller than the minimum transmitted wavelength. %
As the size of the spheres become very small this way, they become challenging to detect in \VWedit[optical ]{RGB-D }sensors simultaneously. 
For this reason, we embed the metal spheres that form a square at the center of comparably larger, \qty{5}{\cm}-diameter styrofoam spheres. 
To support optical stereo technologies, which rely on color features for high-quality depth reconstructions, we colored and textured the styrofoam spheres with random patterns.
Note that, in this way, we induce noise into the radar reconstructions as well, since the material is not completely invisible anymore.
In the next section, we elaborate on how to deal with this noise.
Moreover, we ensure the spatial distance between both, metal and styrofoam spheres is sufficiently large to avoid multi-path effects.
The final calibration target is placed in front of both sensors such that the styrofoam spheres are inside their  respective field of view.

\section{\VWedit{Automatic }Target Detection and Localization}
\label{sec:detection}
The purpose of the target detection and localization stages is to \VWedit{automatically }find the positions of the four metal balls, which are located at the center of the styrofoam spheres.
Since optical and radar sensors operate on different wavelengths, the detection as well as the localization stage is sensor-specific.
During target detection, we aim to find target candidates of point clusters in $\vect{D}_o$ and $\vect{P}_r$, respectively.
In the localization stage, we leverage photometric and spatial constraints to filter these candidates as well as to infer the metal ball locations.

\subsection{Sensor-specific Target Detection}
In the following, we describe the target detection for \VWedit[optical ]{RGB-D }sensors and MIMO radars separately.

\subsubsection{Target Detection in \VWedit[Optical ]{RGB-D }Sensors} 
Since the metal balls are not visible for \VWedit[optical ]{RGB-D }sensors, our method infers their spatial position from the styrofoam sphere centers.
We detect the spheres by utilizing the 2D-3D correspondence of a depth map $\vect{D}_o$ and its respective RGB image $\vect{I}_o$.
Instead of detecting the sphere surface directly, we identify circles in the image plane. While spheres generally map to ellipses, for us this approximation still reliably detected the spheres.
We utilize OpenCV's Circle Hough Transform~\cite{opencv_hough} to find circles in $\vect{D}_o$, in which we clamp the depth values to the near-field range of one meter. 
Note that, in case of stereo vision technologies, in which $\vect{D}_o$ is directly computed from $\vect{I}_o$, this method can also be applied to RGB images instead.
To summarize, our method produces a set of circle candidates, in which four of them are assumed to describe the projected surface of the styrofoam spheres in the image plane.

\subsubsection{Target Detection in MIMO Radars}
The metal balls inside the styrofoam spheres are highly reflective and appear as local maxima in the confidence values of the reconstructed point cloud $\vect{P}_r$.
Compared to the optical domain, reconstructions from radar imaging sensors are more prone to noise and automatic circle detection based on either a projected depth map or confidence map becomes challenging.
In particular, we experienced scattered signals of possibly higher amplitude than the metal balls, which originate from the colored styrofoam spheres that are generally closer to the sensor  (cf. radar top view in \autoref{fig:overview}).
Contrary to related work, which locates corner reflectors based on the assumption of the brightest scatterer, i.e. the target of highest confidence value, we relax this assumption and detect multiple bright scatterers instead.
Given $\vect{P}_r$ and $\vect{I}_r$, we filter out noise and clutter with low confidence based on a decibel threshold $t_{\text{dB}}$.
Next, our method uses greedy non-maximum suppression (GreedyNMS)~\cite{hosang_2017} to find point clusters of high confidence with respect to a Euclidean distance threshold $t_{\text{min}}$. 
In other words, in each iteration we accept the point $\vect{p}_r \in \vect{P}_r$ of highest confidence as a point cluster and reject all other points belonging to the same cluster within a local neighborhood determined by $t_{\text{min}}$.
Moreover, an additional threshold of a maximum Euclidean distance to all previously selected clusters, $t_{\text{max}}$, ensures to select further clusters close to previous ones and avoids the addition of signals from the background. 
After $N \geq 5$ cluster candidates are found, we iteratively assign points, ordered by confidence, to the nearest cluster, until each cluster has $M$ samples.
Based on these samples, we compute the centroid of each cluster.
In this way, we acquire a set ${\mathcal{C}} = \{\vect{c}^k \mid \vect{c}^k \in \mathbb{R}^{ 3} \wedge (k = 1,...,N) \} $ of cluster centers, in which we assume that five of them belong to the metal balls of our calibration target.

\subsection{Sensor-specific Target Localization}
\label{sec:localization}
Analogously to the previous section, the localization of the metal ball centers (from amongst the candidates from the target detection stage) is sensor-specific and described in two separate sections.

\subsubsection{Target Localization in \VWedit[Optical ]{RGB-D }Sensors}
Given the set of circle candidates from the detection stage, we utilize spatial and photometric constraints to find the four circles belonging to the styrofoam spheres.
Similar to GreedyNMS, our method iterates through all detections that are ordered by circle confidence and filters duplicates as well as outliers based on two thresholds for color and size, respectively.
The color threshold filters candidates on the basis of their median color deviation from the ground-truth.
The size threshold discards candidates of significantly deviating radii from already selected circles.
We continue the filtering procedure until four circles are acquired.
Given the intrinsic parameters of $\vect{D}_o$, the circles are projected back into a 3D point cloud. 
To find the relative location of these point clouds with respect to their arrangement on the styrofoam board, we assume that the angle difference between the up-vector of the optical and radar sensor coordinate systems is less than \ang{90}\!.
Based on this assumption, we order the point clouds with respect to the up and right vector of the optical coordinate system.
We denote the resulting set of sphere-shaped point clouds as  $\mathcal{S}_o = \{\vect{S}_o^j \mid \vect{S}_o^j \in \mathbb{R}^{N^j \times 3} \wedge (j = 1,...,4) \} $.

To locate the sphere centers $\Omega_o = (\vect{c}_o^1, ..., \vect{c}_o^4)$ we minimize\TWedit[ the following]{ a} weighted least-squares problem of a sphere equation $L_{\text{data},o}$ with known radius $r$ that is fit to all points $\vect{s}^j_o \in \vect{S}^j_o$:
\begin{align}
  \Omega_{o} &= \underset{\widehat{\Omega}_o}{\text{arg\,min}} \; L_{\text{data},o} \\
  &=  \underset{\widehat{\Omega}_o}{\text{arg\,min}} \sum_{i}^4  \sum_{j}^{N^j} w_{j} \bigl( \lVert \widehat{\vect{c}}^{\;i}_o - \vect{s}^j_o \rVert_2^2  - r^2 \bigr)\;.
\label{eq:optics_data}
\end{align}
Since optical depth sensors suffer from noise in particular at silhouettes, the point-wise error weight $w_j = \langle \vect{n}_j,  \vect{e}_{\text{dir}} \rangle$ describes the range confidence of a point with normal $\vect{n}$ with respect to the sensor's viewing direction $\vect{e}_{\text{dir}}$.
To filter possible outliers, we minimize the energy term multiple times using RANSAC on the point clouds.
During each iteration, we consider a random subset $ \widetilde{\vect{S}}^j_o$ with inlier ratio $k$ and error $e$ (from \autoref{eq:optics_data}) as the current best set $\widetilde{\vect{S}}^j_{o,\text{best}}$ in case the following criterion is fulfilled:
\begin{equation}
	\widetilde{\vect{S}}^j_{o,\text{best}} = \begin{cases}
	\widetilde{\vect{S}}^j_{o} & \lvert k - k_{\text{best}} \rvert > t_{\text{inl}} \lor e < e_{\text{best}}\,, \\
	\widetilde{\vect{S}}^j_{o,\text{best}} & \text{otherwise}\,.
	\label{eq:bestsample}
	\end{cases}
\end{equation}
An inlier ratio threshold,\TWedit[ denoted as]{} $t_{\text{inl}}$, \TWedit[is used as ]{offers }a trade-off parameter between inlier maximization and error minimization.

\subsubsection{Target Localization in MIMO Radars}
To localize the five metal balls %
among the cluster centers detected in $\vect{P}_r$, we use their unique spatial topology established by design of the calibration target.
In this way, it is possible to filter highly ambiguous candidates that may originate from the color-coated styrofoam spheres or other parts of the environment.
Thereby, we utilize the fifth, central metal ball as an anchor point to localize the styrofoam board.
Among all cluster centers  in ${\mathcal{C}}$, our method selects the best subset ${\mathcal{S}_r} = \{\vect{c}_r^k \mid \vect{c}_r^k \in {\mathcal{C}}  \wedge (k = 1,...,4) \}$, together with the anchor point $\vect{c}_{\text{a}, r} \in {\mathcal{C}}$, by minimizing the total weighted energy of:
\begin{equation}
	{\mathcal{S}_r}  = \underset{\widehat{{\mathcal{S}_r}}, \widehat{\vect{c}}_{r,\text{a}} \in {{\mathcal{C}} }}{\text{arg\,min}}\;  L_{\text{data},r} +  \alpha L_{\text{sphere}} + \beta L_{\text{plane}} + \gamma L_{\text{anchor}}\;.
	\label{eq:localization}
\end{equation}
Based on the a priori knowledge that four metal balls lie on a common plane, the term $L_{\text{data,r}}$ minimizes the plane equation parameterized by normal $\vect{n} \in \mathbb{R}^3$ and reference point $\vect{k} \in \mathbb{R}^3$, of which its parameters are estimated along with ${\mathcal{S}_r}$:
\begin{equation}
L_{\text{data},r} = \sum_{\widehat{\vect{c}}_r \in \widehat{{\mathcal{S}_r} }} \bigl\lvert \langle \widehat{\vect{c}}_r - \vect{k},  \vect{n}\rangle  \bigr\rvert\;.
\label{eq:data}
\end{equation}
The regularization term $L_{\text{sphere}}$ enforces each sphere center pair $(\vect{c}^i,\vect{c}^j)$ to be close to the expected spatial distance $d^{i,j}$:
\begin{equation}
	L_{\text{sphere}} = \sum_{i=1}^4 \sum_{j=i+1}^4 \big\lVert \vect{v}_{i,j}\cdot \left( \frac{1}{\lVert \vect{v}_{i,j}\rVert_2} - d^{i,j}\right)\big\rVert_2\;.
	\label{eq:sphere}
\end{equation}
The vector $\vect{v}_{i,j} = f(\vect{\widehat{c}_r}^{\;i}) - f(\vect{\widehat{c}_r}^{\;j})$ describes the relative distance between two spheres $\vect{\widehat{c}_r}^{\;i}, \vect{\widehat{c}_r}^{\;j} \in \widehat{{\mathcal{S}_r} }$, and the function $f(\vect{c}) = \left( \vect{c} - \vect{n} \cdot \langle \vect{c} - \vect{k}, \vect{n} \rangle \right)$ projects the center $\vect{c} \in \mathbb{R}^3$ onto the estimated plane.  In a complementary manner, the term $L_{\text{plane}}$ minimizes the distance $d$ from a center to the styrofoam board, which is localized through the anchor point $\vect{c}_{\text{a},r}$:
\begin{equation}
	L_{\text{plane}} = \sum_{\widehat{\vect{c}}_r \in \widehat{{\mathcal{S}_r} }} \lvert \langle \widehat{\vect{c}}_{\text{a},r} -\widehat{\vect{c}}_r, \vect{n} \rangle - d \rvert\;.
	\label{eq:plane}
\end{equation}
Lastly, the term $L_{\text{anchor}}$ ensures that the anchor point $\vect{c}_{\text{a},r}$ lies in the center of the square sphere arrangement:
\begin{equation}
L_{\text{anchor}} = \big\lVert f(\widehat{\vect{c}}_{\text{a},r}) - \sum_{\widehat{\vect{c}}_r \in \widehat{{\mathcal{S}_r} }}  \frac{1}{4} \widehat{\vect{c}}_r \big\rVert_2\;.
\label{eq:anchor}
\end{equation}
To find  ${\mathcal{S}_r}$, our method tries $\binom{N}{5}$ combinations to sample five cluster centers ${\widehat{\mathcal{S}}_r}$  in each iteration.
Since four of these centers should lie on a common plane, we test all possible $\binom{5}{4}$ combinations to find the plane $(\vect{k}, \vect{n})$, which minimizes~\autoref{eq:data}.
Similar to optical localization, the four metal ball candidates of the current sample are ordered with respect to the up and right vector of the sensor coordinate system.
Next, the current sample is evaluated in terms of its inlier ratio, using ~\autoref{eq:data}, and the error function given in ~\autoref{eq:localization}.
Lastly, we compute~\autoref{eq:bestsample} to determine, whether the sample ${\widehat{\mathcal{S}}_r}$ is better than the current best random subset.
To show the necessity of the additional spatial constraints in ~\autoref{eq:localization}, an ablation study is performed in~\autoref{sec:eval}.

\section{\VWedit{Automatic }Spatial Registration and Refinement}
Based on the previously localized metal balls, the final stage \VWedit{automatically }computes the relative rigid transformation between the coordinate systems of a sensor pair through spatial registration.
Moreover, we support an optional refinement stage, utilizing a second object of simpler geometry to establish a significantly higher amount of correspondence pairs.
As we will show, this optional stage confirms our method's accuracy but does not significantly improve results.

\subsection{Calibration Parameter Estimation}
Based on their spatial location on the styrofoam board, we find ordered pairs $(\vect{c}^i_r, \vect{c}^i_o)$ of sphere centers in the radar and optical sensor coordinate system, respectively. 
We assume that both sensor coordinate systems are metrical and use a priori knowledge about their factory settings to determine the uniform scale matrix $\vect{S} \in \mathbb{R}^{3 \times 3}$ from the optical to the radar coordinate units.
Then, we solve for the optimal rotation $\vect{R} \in \mathbb{R}^{3 \times 3}$ and translation $\vect{t} \in \mathbb{R}^3$ from the optical coordinate system to the radar coordinate system by minimizing their root mean square error. 
We use the closed-form solution of Kabsch~\cite{kabsch_1976} to acquire $\vect{R}$ and $\vect{t}$ as follows:
\begin{align}
	\vect{H} &= \vect{\bar{C}}_r^T \cdot \vect{S} \cdot \vect{\bar{C}}_o \stackrel{\text{\scriptsize SVD}}{\;=\joinrel=\;} \vect{U} \cdot \Sigma \cdot \vect{V}^T \\
	\vect{R} &= \vect{U} \vect{V}^T \\
	\vect{t} &= \bar{\vect{c}}_r - \vect{R} \cdot \vect{S} \cdot \bar{\vect{c}}_o 
\end{align}
The mean of the four sphere centers in sensor domain $*$ is denoted as $\bar{\vect{c}}_*$. The data matrices $\vect{\bar{C}}_r, \vect{\bar{C}}_o \in \mathbb{R}^{4\times 3}$ contain the mean-centered spheres in pairwise order. 
Since $\vect{S}$ is computed a priori, $\Sigma$ is expected to be close to the identity matrix.

\subsection{Calibration Refinement for Evaluation}
As part of our evaluation (\autoref{sec:eval}), we employ an additional refinement stage as a method to assess potential calibration errors arising from uncertainties with respect to the sensor range.
In this stage, the target for establishing correspondences is a simple, textured metal plate mounted on a styrofoam board for an upright standing. 
The amount of received signal from a planar surface is strongly view dependent for a MIMO radar such that the plate has to be placed parallel to the antenna aperture.
Since it can be assumed that the first estimate of $\vect{R}$ and $\vect{t}$ is accurate enough, we compute correspondences via a projective mapping.
Given the intrinsic parameters of $\vect{D}_o$, we transform all points in $\vect{P}_r$ to the optical image plane and establish correspondence pairs based on points sharing the same pixel coordinate.
Next, we repeatedly solve for $\vect{R}$ and $\vect{t}$ using the Kabsch algorithm in combination with RANSAC, in which we randomly sample correspondence pairs to minimize~\autoref{eq:bestsample}.

\section{Evaluation}
\label{sec:eval}
In this section, we describe the evaluation setup as well as the quality assessments of our calibration.

\begin{figure}[b]
	\centering
	\includegraphics[width=0.75\linewidth]{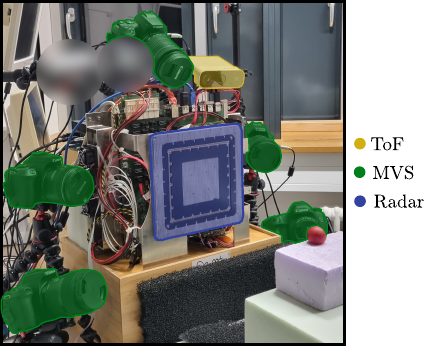}
	\caption{\label{fig:setup}
		Our setup consists of an imaging MIMO radar, a Kinect Azure camera (ToF) and five DSLR cameras (MVS).}
\end{figure}

\subsection{Evaluation Setup}
The MIMO radar is a submodule of an Automotive Radome Tester provided by Rohde \& Schwarz~\cite{rs_qar}.
Its virtual aperture consists of $94 \times 94$ transmitting and receiving antennas, arranged on a square frame.
The signal form is \textit{stepped frequency continuous wave} within a frequency range from \qty{72}{\GHz} to \qty{82}{\GHz} with 128 frequency steps~\cite{braeunig_2023}.
To acquire $\vect{P}_r$ and $\vect{I}_r$, we make use of a state-of-the-art reconstruction method for millimeter wave imaging, which is known as back-projection and described further in ~\cite{braeunig_2023, sherif_2021}. 
$\vect{P}_r$ is reconstructed in a range between 20--\qty{65}{\cm}.
To demonstrate that our calibration target can be used for various optical technologies, we employ two distinct depth sensing methods:
 \textit{amplitude modulated continuous wave} time-of-flight (ToF) using the Microsoft Kinect Azure~\cite{microsoft_kinect} camera, and multi-view stereo (MVS) using five Canon DSLR cameras with $>\qty{24}{MP}$ resolution.
The setup is depicted in~\autoref{fig:setup}.
We evaluate the calibration within a constrained environment using styrofoam as a rest table, a black screen made of fabric, and absorbers behind.
Our experiments are divided into two scenarios: first, we record the calibration target, in a constrained position, followed by a more natural capture process.
With respect to the former, we place the calibration target on a plastic turntable such that it is centered in the radar coordinate system.
The anchor metal ball represents the center of its rotation.
We place the turntable at \qtylist{30;40;50}{\cm} distance to the radar, respectively, and rotate the calibration target between $[\ang{-20}, \ang{20}]$ in steps of~\ang{5}.
To simulate a natural capture process, we relocate the calibration target multiple times within a range of 30--\qty{50}{\cm} distance such that it is roughly centered by eye with respect to the radar.
In this way, we acquire 40 different calibration target captures.
Lastly, we record the refinement object once at \qty{30}{\cm} distance.
We use this object in our experiments only if explicitly noted.

In our calibration pipeline, we set the parameters $t_{\text{dB}} = 15$, $t_{\text{min}} = \qty{2}{\cm}$, $t_{\text{max}} = \qty{30}{\cm}$, $t_{inl} = 0.05$, $N=20$, $M=7$, $\alpha = 2$, $\beta = 2$, and $\gamma = 4$.
Furthermore, we set the maximum number of RANSAC iterations in the optical localization stage to 1000 and in the optional refinement stage to 100.

\subsection{Results}
We show the efficiency of our method in three analyses.
First, we evaluate the performance with respect to different orientations and distances of the calibration target.
Second, we perform an ablation study to demonstrate the importance of the target design together with the utilization of spatial constraints in the radar domain.
Lastly, we show qualitative results with respect to three captured objects.
We captured objects of distinctive geometry and color, on which we measure the calibration error: a metal disk (at \qty{30}{\cm} and \qty{40}{\cm} distance to the MIMO radar), a symbol cut out from cardboard (at \qty{30}{\cm}), and a 3D-printed hand model coated in metal lacquer (at \qty{30}{\cm}).
Our primary metric is the Chamfer distance $C$ between object points $\vect{p}_o \in \vect{P}_o$ of an optical sensor and points $\vect{p}_r \in \vect{P}_r$ of the MIMO radar:
\begin{equation}
	C = \frac{1}{2} \text{RMSE}( \vect{P}_o,  \vect{P}_r) + \frac{1}{2} \text{RMSE}( \vect{P}_r,  \vect{P}_o)
\end{equation}
The root mean square error is calculated on the Euclidean norm per point pair, established based on nearest Euclidean distance.
In the following, we will address each experiment in more detail.

\subsubsection{View- and Distance-dependent Calibration}

\begin{figure}[b]
	\centering
	\includegraphics[width=0.9\linewidth]{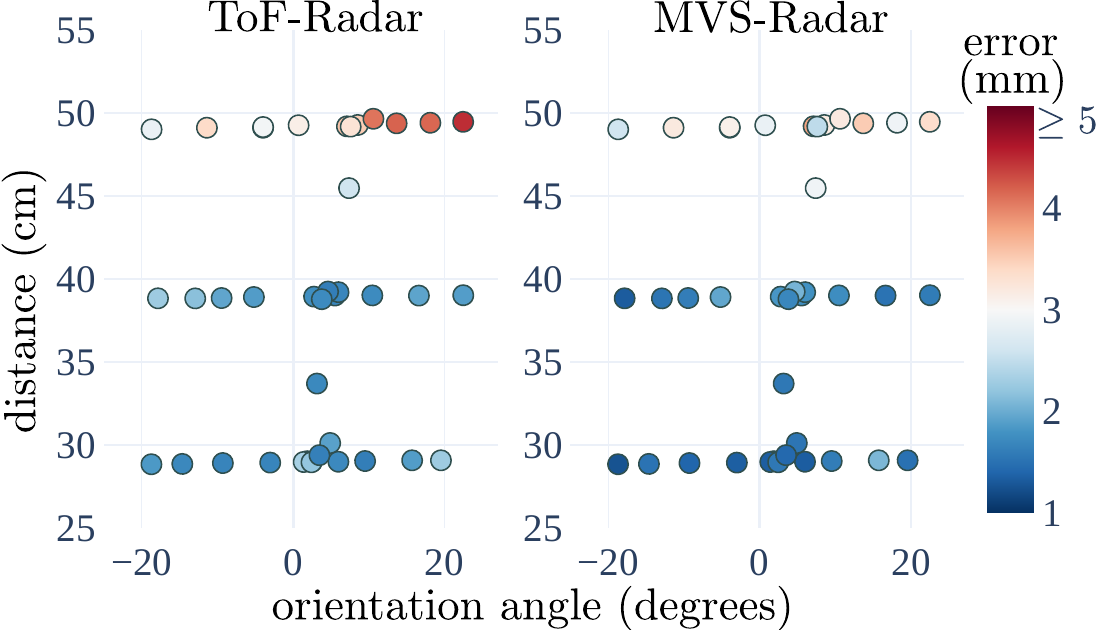}
	\caption{The Chamfer distance for the ToF-radar and MVS-radar sensor pair, respectively. We plot the target angles and distances from the estimated plane in the radar coordinate system during sphere localization.}
	\label{fig:results_angle}
\end{figure}

\begin{figure}[t]
	\centering
	\includegraphics[width=1\linewidth]{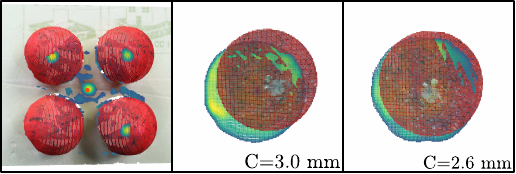}
	\caption{\label{fig:results_distance}%
          MVS-radar calibration of a target at \qty{50}{\cm} distance
          (\textit{left}). The anchor point (orange) approximately
          aligns with the radar signal. The error of a disk at
          \qty{30}{\cm} (\textit{middle}) and \qty{40}{\cm}
          (\textit{right}) decreases with its distance to the target.}
\end{figure}
To demonstrate that our calibration target can be placed in front of a MIMO radar without giving a considerable amount of attention to its precise placement, we assess the calibration accuracy with respect to multiple orientation angles and distances.
More specifically, we calculate the Chamfer distance of the cardboard symbol that was captured at \qty{30}{\cm} distance to the MIMO radar.
Results are shown for both, the ToF-radar and MVS-radar sensor pair in~\autoref{fig:results_angle}.\VWedit{Additionally, for both pairs, we ran the calibration 20 times for the same target at (30~cm,~$0^\circ$), yielding a standard deviation of $\pm0.004^\circ$ for the average rotation and $\pm0.17$~mm for the average translation.}

For both pairs, our method works best within all recorded orientation angles and a distance between 30--\qty{40}{\cm}. 
Within this range, all samples of the ToF-radar and MVS-radar calibration have an average Chamfer distance of \qty{1.69}{\mm} and \qty{1.72}{\mm}, respectively, regardless of the target orientation.
The average error increases by \qty{0.53}{\mm} and \qty{0.54}{\mm} when including the results at $\geq$\qty{50}{\cm} distance.
Upon further investigation, we observe that the Chamfer distance significantly depends on the spatial distance between the calibration location and the location of the evaluation object.
In~\autoref{fig:results_distance}, the results for a calibration target, captured at \qty{50}{\cm} distance, are depicted with respect to the metal disk, placed at \qty{30}{\cm} and \qty{40}{\cm} distance, respectively.
This example illustrates that the error decreases with the distance between the calibration target location and the evaluation object.
Hence, we conclude that the calibration is only valid within a specific range due to perspective distortion and systematic range errors of the \VWedit[optical ]{RGB-D }sensors.
Moreover, the ToF camera exhibits a comparably large Chamfer distance at \qty{50}{\cm} distance and orientations $>\ang{5}$ when compared to the MVS-radar calibration. 
Assessed from the average calibration parameters, the ToF coordinate system has a spatial offset of $\{\qty{5}{\cm}, \qty{17}{\cm}, \qty{15}{\cm}\}$ and $\{\ang{12}, \ang{15}, \ang{20}\}$ with respect to the (right, up, direction) vector triple of the radar coordinate system.
As a consequence, the angle between the target plane and the ToF view direction is $+\ang{16}$ larger than in the radar coordinate system, such that results at $\ang{22}$ in~\autoref{fig:results_angle} have an orientation of $\ang{38}$ in the ToF coordinate system.
To summarize, the application-dependent working distance has to be considered during target calibration.
Inside this working distance, our approach achieves millimeter accuracy far below the radar wavelength (\qty{3.7}{\mm}) and the random noise distribution ($\leq \qty{17}{\mm}$) of the ToF camera.
Since our calibration stays accurate, regardless of the target orientation in our experiments, we conclude that it is not required to balance the target precisely in front of the sensors.

\begin{figure*}[t!]
  \centering
  \includegraphics[width=0.95\linewidth]{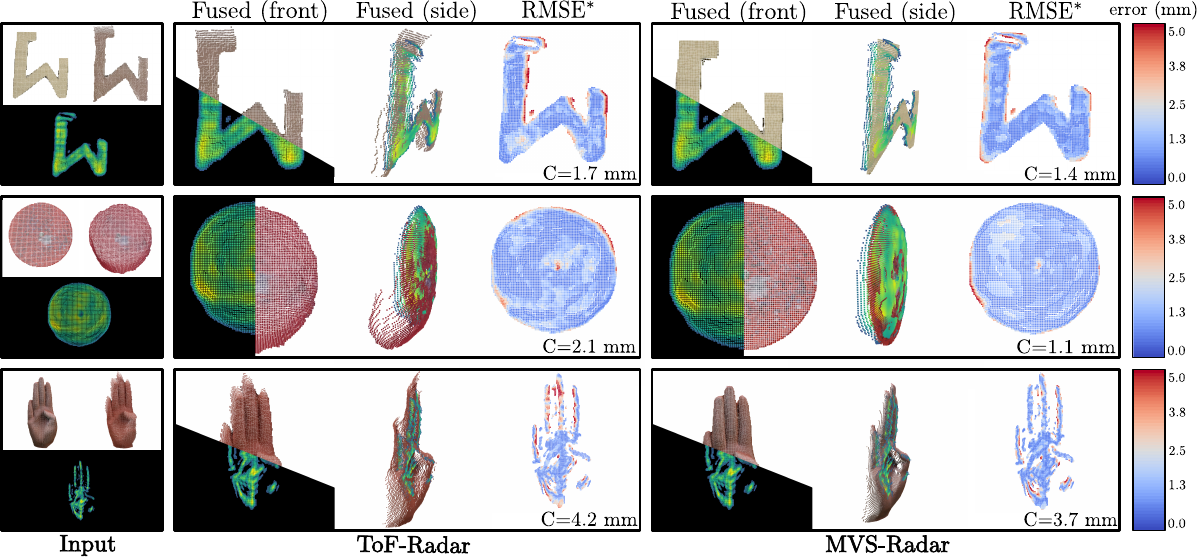}
  \caption{\label{fig:results_qualitative}%
    We show qualitative results for a cardboard symbol, a disk, and a 3D printed hand, respectively. The Chamfer distance $C$ is indicated in the bottom of the error visualization. \TWedit[$*$ ]{RMSE$^\ast$ }denotes the point-wise RMSE between a radar point and the nearest point of Euclidean distance from the optical sensor.}
\end{figure*}
\begin{figure}[b]
  \vspace*{-2ex}\centering
  \includegraphics[width=0.9\linewidth]{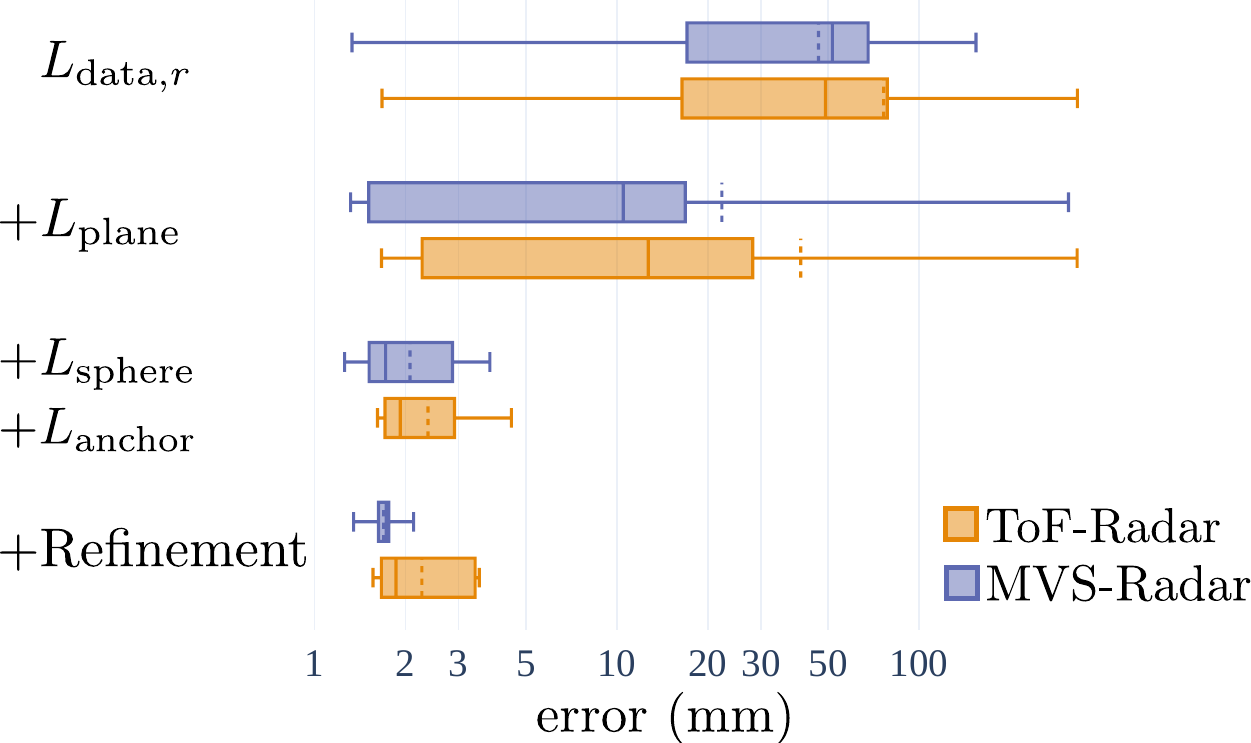}
  \caption{\label{fig:results_ablation}%
    We vary the spatial constraints during radar sphere localization to simulate the availability of target elements added during the design. We also assess the calibration quality after the refinement stage. The median and the mean of each box plot are marked as solid and dashed lines, respectively.}
\end{figure}

\subsubsection{Ablation Study}
We demonstrate the necessity of the choices made during calibration target design as well as the utilization of these choices through spatial constraints in an ablation study.
The results with respect to the Chamfer distance of the cardboard symbol are given in~\autoref{fig:results_ablation}.
Since our main contribution lies in a near-field calibration of an imaging MIMO radar, the focus of this study is on target localization in the radar domain.
Without any systematic spatial arrangement of the spheres, the only term that can be applied during localization is $L_{\text{data},r}$ and the resulting calibration error is within \qty{5}{\cm} on average.
By arranging the spheres in a square, the additional regularization term $L_{\text{sphere}}$ can be employed, which decreases the calibration error by \qty{4}{\cm} on average.
Lastly, the anchor point that is mounted in the center of the square on the styrofoam board leads to another error decrease by \qty{8}{\mm} through the utilization of the regularization terms $L_\text{plane}$ and $L_\text{anchor}$.
To conclude, the results demonstrate that the square arrangement with five metal balls is necessary to achieve a calibration quality below \qty{2}{\mm}.
So far, our calibration is single-shot, which means it only requires one capture of the calibration target\VWedit{ and, thus, little capture effort}.
We further show results of a second capture, in which we record the metal plate and perform the refinement stage.
While the median of both, the ToF-radar (\qty{1.86}{\mm}) and MVS-radar (\qty{1.72}{\mm}) calibrations are similar to calibration without refinement (\qty{1.92}{\mm} and \qty{1.72}{\mm}), we observe a common decrease in the error variance, together with the mean. 
This decrease is due to the improving calibration results specifically at $\geq \qty{45}{\cm}$ distance, since the refinement target is placed at \qty{30}{\cm} and, therefore, is able to correct the misalignment arising from the spatial distance between the position of calibration target and the cardboard symbol.
However, we argue that in these cases it would have been simpler to place the calibration target at \qty{30}{\cm} in the first place.
In summary, we demonstrate that each of our target design choices is necessary and the calibration can not be further improved by a second capture.

\subsubsection{Qualitative Results}
In the last experiment, the calibration accuracy of a target placed at \qty{30}{\cm} distance is assessed in qualitative results using the three additional captured objects of distinctive geometry complexity and distance.
In~\autoref{fig:results_qualitative}, we estimate the Chamfer distance and the point-wise RMSE from a radar point to the nearest point, and show results for the cardboard, the metal disk and the hand, each recorded at a distance of \qty{30}{\cm}.
For all objects, the RMSE is primarily below \qty{2}{\mm}.
Its distribution is geometry- and sensor-specific.
Moreover, the ToF-radar alignment results in higher point-wise errors due to the fact that active ToF cameras exhibit more noise than high-resolution passive MVS algorithms.
The cardboard symbol in the first row has the best average alignment quality for both, ToF-radar and MVS-radar calibrations.
The disk in the second row exhibits a comparably higher Chamfer distance due to flying pixel artifacts in the ToF camera.
Lastly, the hand in the third row of~\autoref{fig:results_qualitative} demonstrates the huge domain gap between reconstructions of optical and radar sensors.
For complex geometries, most of the signals do not return back to the MIMO radar, which underlines the importance of careful calibration target design.
In summary, the qualitative results further corroborate the accuracy of our calibration method and offer interesting findings in terms of the domain-specific sensor characteristics.

\section{Conclusion}
We presented a novel calibration method for optical technologies in combination with an imaging MIMO radar in the near field within centimeter range.
Considering the large domain gap between the two frequency domains, we designed a suitable calibration target that consists of four textured styrofoam and five metal balls, arranged at the corners and the center of a square.
Given a capture of this target, our method detects circles in the optical domain, and clusters points of high target confidence in the radar domain.
Due to careful design of the target's spatial arrangement, we utilized photometric as well as spatial constraints to detect and localize the four metal balls within each sensor coordinate system.
Finally, we compute the calibration parameters through spatial registration of these balls and propose to assess the alignment quality in an optional refinement stage.
In the evaluation, we demonstrate the effectiveness of our target design, eliminating the need for careful positioning in front of a sensor, and show the importance of the spatial arrangement.
In summary, our calibration target is  single-capture, user-friendly with respect to its placement, and yields millimeter accuracy up to a remaining error of less than \qty{2}{\mm}, which is considerably small such that it may originate from sensor noise itself.

\section*{Acknowledgement}
The authors would like to thank the Rohde \& Schwarz GmbH \& Co. KG (Munich, Germany) for providing the radar imaging devices. This work was funded by the Deutsche Forschungsgemeinschaft (DFG, German Research Foundation) – SFB 1483 – Project-ID 442419336, EmpkinS. 

\printbibliography

\end{document}